\documentclass[conference]{ieeeconf}
\IEEEoverridecommandlockouts 
\usepackage{cite}
\usepackage{amsmath,amssymb,amsfonts}
\usepackage{algorithmic}
\usepackage{textcomp}
\usepackage{xcolor}
\usepackage{booktabs} 
\usepackage{blindtext, graphicx}
\PassOptionsToPackage{greek,english}{babel}
\usepackage{soul,color}
\usepackage{comment} 
\usepackage{kantlipsum}
\usepackage{multicol}
\usepackage{array}
\usepackage{tabulary}
\usepackage{chngcntr}
\usepackage{multirow}
\newcolumntype{K}[1]{>{\centering\arraybackslash}p{#1}}

\usepackage{float}
\usepackage[linesnumbered,ruled,vlined]{algorithm2e}
\usepackage{bm}
\newcolumntype{M}[1]{>{\centering\arraybackslash}m{#1}}

\usepackage{cleveref}
\title{
ReLearn: A Robust Machine Learning Framework in Presence of Missing Data for Multimodal Stress Detection from Physiological Signals*
}

\author{Arman Iranfar, Adriana Arza, and David Atienza$^{1}$
\thanks{*This work has been partially supported by the ML-Edge Swiss National Science Foundation (NSF) Research project (GA No. 200020182009/1), in part by the DeepHealth H2020 Project (GA No. 825111), and by the ONR-G through Award Grant No. N62909-20-1-2063.}
\thanks{$^{1}$A. Iranfar, A. Arza, and D. Atienza are   with the Embedded Systems Laboratory of Swiss Federal Institute of Technology Lausanne,    Switzerland.
        {\tt\small \{arman.iranfar, adriana.arza, david.atienza\}@epfl.ch}}%
}

\begin{document}

\maketitle
\thispagestyle{empty}
\pagestyle{empty}

\begin{abstract}
Continuous and multimodal stress detection has been performed recently through wearable devices and machine learning algorithms. However, a well-known and important challenge of working on physiological signals recorded by conventional monitoring devices is missing data due to sensors insufficient contact and interference by other equipment. This challenge becomes more problematic when the user/patient is mentally or physically active or stressed because of more frequent conscious or subconscious movements. 
In this paper, we propose ReLearn, a robust machine learning framework for stress detection from biomarkers extracted from multimodal physiological signals. ReLearn effectively copes with missing data and outliers both at training and inference phases. ReLearn, composed of machine learning models for feature selection, outlier detection, data imputation, and classification, allows us to classify all samples, including those with missing values at inference. In particular, according to our experiments and stress database, while by discarding all missing data, as a simplistic yet common approach, no prediction can be made for 34\% of the data at inference, our approach can achieve accurate predictions, as high as 78\%, for missing samples. 
Also, our experiments show that the proposed framework obtains a cross-validation accuracy of 86.8\% even if more than 50\% of samples within the features are missing. 
\end{abstract}

\section{INTRODUCTION}
\bstctlcite{IEEEexample:BSTcontrol}
Stress, as a global issue of modern societies, increases the risk of several health pathologies, such as, heart diseases, depression, and sleep disorders \cite{cohen2007psychological}. 
Continuous and multimodal stress detection and recognition have been realized through wearable devices and embedded machine-learning algorithms, using stress biomarkers extracted from different physiological signals, such as photoplethysmography (PPG), respiration (RSP), electrodermal activity (EDA), electrocardiogram (ECG), and skin temperature \cite{Smets2018, montesinos2019multi, arza2019measuring,Parent2020PASS:Research}. 

A well-known challenge of working on physiological signals recorded by conventional monitoring devices~\cite{zhang2018cgmanalyzer} is the presence of missing data due to sensors or electrodes insufficient contact and user's motion, as well as interference by other equipment~\cite{e21030274}. This situation deteriorates if the user is physically or mentally active or stressed as a consequence of more frequent conscious or subconscious movements. In general, 
handling missing data and outliers is of paramount importance when solving real-life classification and regression problems through pattern recognition techniques \cite{ding2015missing}. On one hand, classification and regression models should be fit offline with a complete and flawless training dataset, i.e., without missing values or outliers. On the other hand, these models, when online, should be still able to provide accurate enough predictions even in presence of missing data and outliers. Therefore, it is vital to have a machine learning framework sufficiently robust to the effect of incomplete data, especially for biomedical applications where prediction accuracy directly or indirectly affects human life quality. 

To address such issues in general-purpose applications, researchers have modified the traditional pattern recognition techniques to consider outliers and missing data \cite{pygmmis,chen2009robust}. 
However, these techniques lie in certain assumptions, excluding the nature of physiological signal recording and biomarker extraction from wearable devices, where it is possible to encounter a long missing segment of data \cite{moody2010physionet}. 

Although several works \cite{jadhav2010artificial, bai2020fatigue} in the biomedical application domain have considered the impact of incomplete data, they are very simplistic, providing inaccurate predictions. Moreover, to the best of our knowledge, stress detection from physiological signals in presence of missing data has not been taken into account in the literature. 
Therefore, a comprehensive framework that deals with missing data in stress biomarkers throughout its whole pipeline needs to be addressed. 

In this work, we propose \textbf{ReLearn}, a robust machine learning framework for stress detection from biomarkers extracted from multiple physiological signals that effectively copes with missing data both at training and inference phases. In particular, the training dataset composed of stress biomarkers with missing values flows into a pipeline of feature selection, data imputation, and outlier detection machine learning algorithms, such that the classifier can be fed with a complete training dataset. Then, at inference, unseen data including the missing values are first passed through the trained models obtained from the feature selection, data imputation, and outlier detection algorithms. Finally, the imputed data are used by the classifier to make predictions.

Our main contributions in this work are as follows:
\begin{itemize}
    \item We propose a novel machine learning framework for multimodal stress detection from physiological signals, which is robust to missing data and outliers.
    \item  According to our experiments and stress database, while by discarding all missing data, as a simplistic yet common approach, no prediction can be made for 34\% of the data at inference, our approach is able to achieve accurate predictions, as high as 78\%, for missing samples.
    \item Our experiments show that the proposed framework obtains a cross-validation accuracy of $86.8\%$ even if more than $50\%$ of samples within the features are missing. 
\end{itemize}

\section{RELATED WORK}
Several works in the literature deal with outliers, noise, and missing data in physiological signals. 
In this context, \cite{bai2020fatigue} only considers parts of signals with valid values, discarding all missing data (NAN values). Nonetheless, throwing away part of the data makes it impossible to have any prediction about the patient's or user's condition. 
Authors in \cite{jadhav2010artificial} replace missing values with the closest valid values of the corresponding point. However, such an approach only suits situations where missing data occur infrequently. Also, if more than a few successive data points are missing, filling this gap with the last valid value is insufficient and the outcome could be misleading. 

Besides such simplistic approaches, there are several more complex methods addressing missing and noisy data for different applications. In particular, 
\cite{langley2010estimation} uses a reference channel to substitute the missing data of physiological time series. Since this method works directly on the raw biosignals, it is not suitable for machine learning approaches, where rather than the raw signal, the extracted features are used as the input data. \cite{afrin2018simultaneous} reconstructs the missing leads of a 12-lead ECG signal from a single-lead ECG signal by using the Random Forest algorithm. Nevertheless, this work does not address how to cope with missing data for other physiological signals, such as PPG, RSP, etc. 
A Singular Value Decomposition (SVD) analysis of outlier detection and imputation of missing data is presented by \cite{liu2003robust} for DNA microarrays. Similar to \cite{afrin2018simultaneous}, the work of \cite{liu2003robust} is application-specific and cannot be used as a general solution. 
Adaptive filtering is leveraged by \cite{hartmann2010reconstruction} to predict a 30-second segment of missing cardiovascular signals. This approach, however, falls short if long enough segments of signals without any missing data cannot be found before the missing segment. A high dimensional Gaussian Mixture Model (GMM) is built by \cite{ding2015missing} to address classification problems in high-dimensional samples with missing values. Although this approach shows promising results for surface electromyography (sEMG) signals, it does not provide a holistic solution for multimodal physiological signals. Similarly, authors in \cite{ding2019adaptive} propose an adaptive incremental hybrid classifier to alleviate the impact of outliers in myoelectric pattern recognition. 
A more general-purpose framework is proposed by \cite{kiasari2017novel} where an iterative algorithm is used for classifying data with a missing feature. Although the proposed algorithm has been tested on different datasets and applications, it neither considers nor examines the effect of outliers on the classification task. 

Our proposed machine learning framework, in contrast to state-of-the-arts, provides a comprehensive solution that can work on arbitrary physiological signals while addressing missing data and outliers.

\section{PROPOSED FRAMEWORK}
\begin{figure}
    \centering
    \includegraphics[scale=0.8]{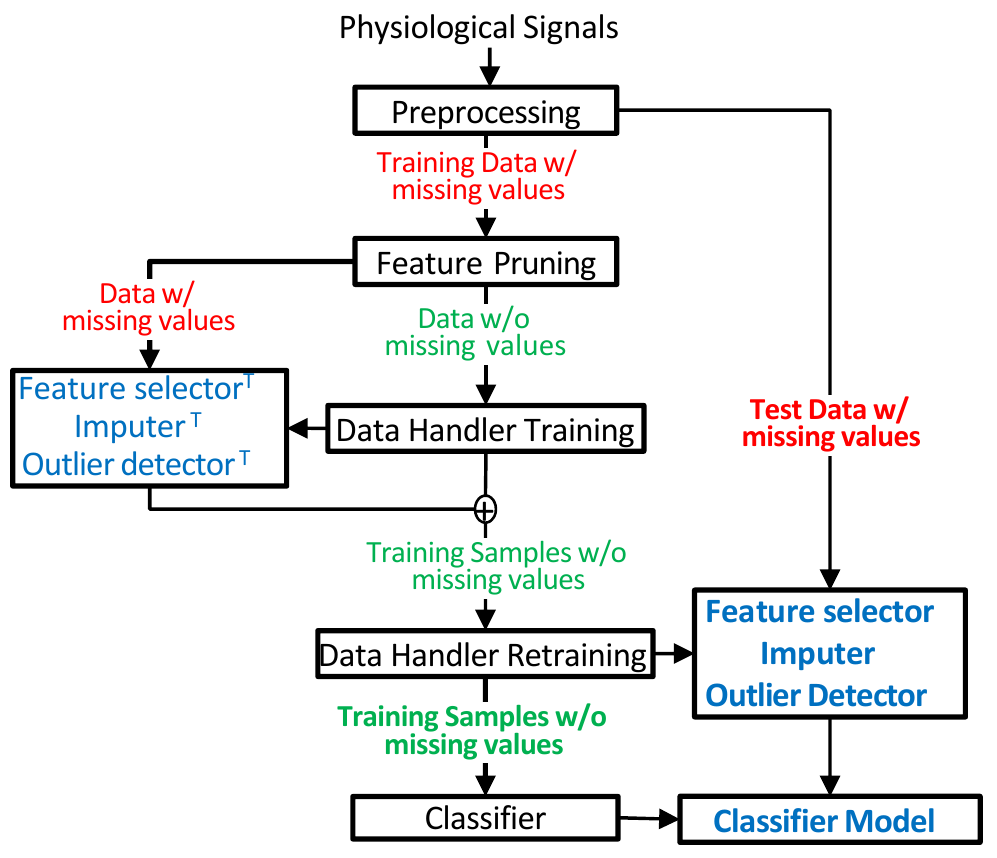}
    \caption{Overall view of proposed framework for multimodal stress detection in presence of missing data}
    \label{fig:overall_framework}
    \vspace{-0.5cm}
\end{figure}

In this work, we design a machine learning framework for stress detection from multimodal physiological signals, which can cope with missing values and outliers at both training and inference time. 
Fig. \ref{fig:overall_framework} shows an overall view of the proposed framework. First, multimodal physiological signals are preprocessed to extract input features (i.e., stress biomarkers) of machine learning algorithms and create the training and testing datasets, both including several samples with missing values. Second, we propose to prune the features, i.e., to exclude those whose ratio of missing value over all samples of the training data is above a particular predefined threshold. So that we reject those features that are prone to missing values, hence, the more affected by the noise and artifacts. However, since important physiological information could be loosed, the trade-off between looser/weaker pruning and cross-validation accuracy is assessed in Section \ref{sec:threshold}.

Then, the training samples without missing values are used to train the \textit{Data Handler}, where machine learning models for feature selection, outlier detection, and data imputation are trained. Those models are used to clean and impute the training data with missing values, resulting in training samples without any missing values. 
Thereafter, the data handler is retrained with the complete and enhanced data without missing values to create our ultimate feature selector, data imputer, and outlier detector. 
Afterward, the machine learning classifier is trained with the training data without any missing values and outliers. 
At inference time, our ultimate retrained data handler (feature selector, outlier detector, and data imputer) 
is used to deal with missing values and outliers of unseen testing data prior to the classification. Finally, the trained classifier is employed to provide real-time predictions.

Throughout this framework, we apply cross-validation (CV) with balanced accuracy score \cite{brodersen2010balanced} for training the feature selection algorithm, as well as the classifier. For this purpose, we apply a group K-fold cross-validation, with K=10, where for each iteration of the cross-validation, a couple of the existing groups (in our case, subjects), are kept for validation, while the training is performed for the rest of the groups. The group K-fold cross-validation is desirable since it can avoid overfitting, particularly, when the samples at inference most likely come from completely different subjects.
In what follows, we detail each stage of the proposed framework.

\subsection{Signal Preprocessing}
\label{sec:preprocesssing}
The data-flow from the raw physiological signals to training and testing datasets is shown in Fig. \ref{fig:dataflow}. The first step is the signal preprocessing, wherefrom each physiological signal a set of features that capture the subject's physiological stress response is extracted in segmentation widows of 60s. 

First, the raw signals are filtered to remove noise and artifacts. Second the filtered signals are delineated to obtain the primary parameters as in \cite{arza2019measuring, DellAgnola2020} and \cite{montesinos2019multi}, which are shown in Fig. \ref{Fig:signals}. In this step, for each parameter, several data quality policies are applied mainly based on physiological expected values (e.g., heart rate from 30-180 bpm) and previous samples trend (i.e., within mean, median, and standard deviation values ranges of the last 3 to 5 samples). Besides, each delineation algorithm also rejects noisy signal-segments based on the expected physiological signal shape. 

Next, 94 physiological features in the time and frequency domain are extracted from the parameters time series in segmentation windows of 60s, as described in our previous works \cite{montesinos2019multi, Masinelli2020}. Here again, several policies are applied for the missing data on the parameters time series when extracting the features on each segmentation window to ensure the characterization of the physiological response on the sample. For instance, frequency features are only computed if we have more than $98\%$ of the data; heart cycle-based features (on ECG and PPG signals) are valid if more than 10 heartbeats are delineated in a segmentation window; similarly with the respiration cycles, more than 5 cycles. In the case of the EDA signal, its features return a missing value if less than $80\%$ of the data is available.
These features are described as follows: 
\subsubsection {EDA} The EDA signal is divided into two main components: Skin Conductance Level (SCL) and  Skin Conductance Response (SCR) as the driver phasic signal \cite{Benedek2010}. Then, the gradient and mean of the SCL, as well as the SCR power are obtained. 
\subsubsection {RSP} 
We compute respiration period ($RSP_{PRD}$), duration of air inhaled ($INS_{time}$), and exhaled ($EXP_{time}$), and the ratio of inhalation to exhalation duration, from which statistical features are extracted. In the frequency domain, we compute the power and normalized band power of the segmented signal in different frequency bands. 
Moreover, for each window of analysis, we applied the method proposed in \cite{Hernando2016} to compute the estimated respiratory frequency, the largest peak power, the total power, and the normalized respiratory peak power. 
\subsubsection {ECG} From ECG, the time intervals between two consecutive R peaks (RR) are obtained. From the RR interval, several time and frequency domain features are extracted based on the Heart Rate Variability (HRV) analysis \cite{TaskForceo1996}. Non-linear features are also extracted from the Poincaré plot indicating vagal and sympathetic function. 
\subsubsection {PPG} 
\begin{figure}[t]
	\centering
	\includegraphics[width=0.9\columnwidth]{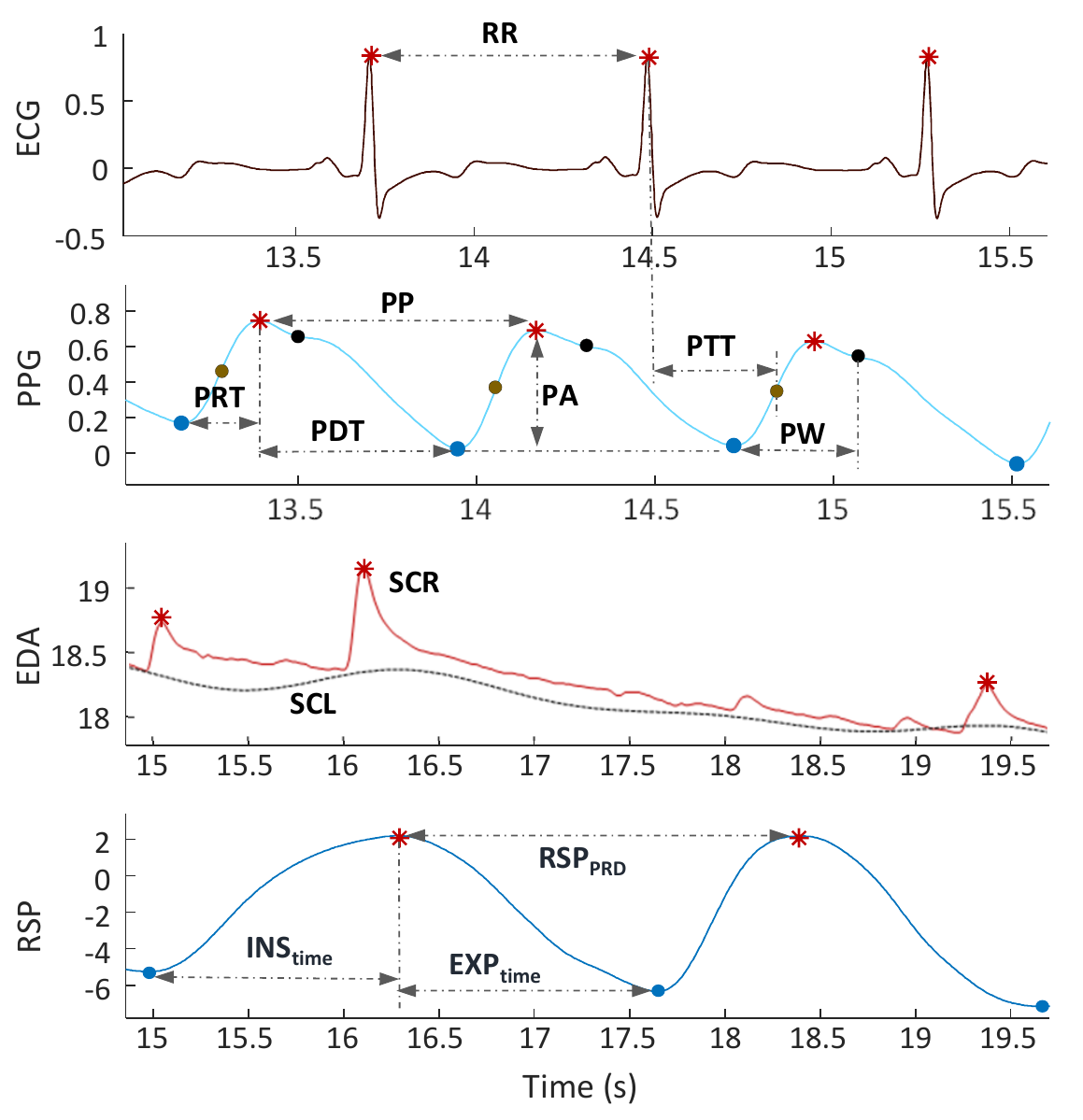}
	\caption{Biomarkers extracted from electrodermal activity (EDA), respiration (RSP), electrocardiogram (ECG) and photoplethysmography (PPG) signals. }
	\label{Fig:signals}
	\vspace{-.5cm}
\end{figure}

Several parameters are computed as represented in Fig. \ref{Fig:signals}: pulse period (PP), pulse wave rising time (PRT), pulse wave decreasing time (PDT), pulse width until reflected wave (PW), pulse amplitude (PA), and the slope of the pulse (k) defined as the slope transit time between 1/4 and 3/4 of the PA divided by their difference in amplitude. PP interval features are equal to the aforementioned for RR intervals. 


\subsection{Feature Pruning }
\begin{figure*}
    \centering
    \includegraphics[scale=0.8]{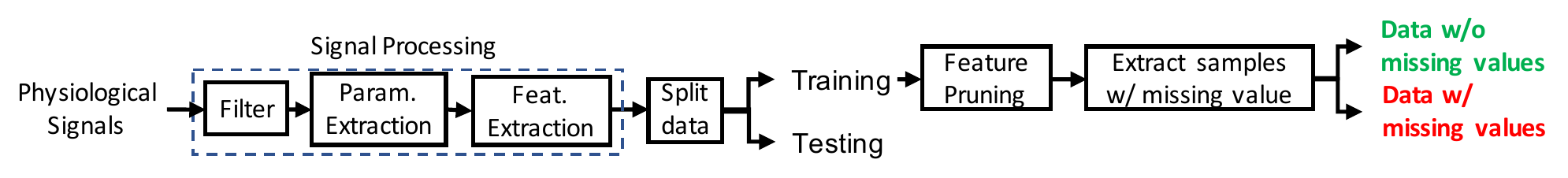}
    \caption{Preprocessing stage of our proposed framework}
    \label{fig:dataflow}
\end{figure*}

After feature extraction, we randomly select $70\%$ of the subjects for training, i.e., 66 and 29 subjects' data, respectively, for training and testing, while making sure no subject's data in the training dataset appears in the testing datasets, cf. Fig. \ref{fig:dataflow}. We split the data in a stratified fashion such that the same proportion of class labels exists in both datasets. 

Our prepossessing stage by applying the missing/noisy data policies results in a dataset with reliable samples but also missing ones. Therefore, 
we only consider those features of training data that have values for at least more than a predefined percentage of the samples on the training set on the pruning step. The lower the threshold is, the fewer features prone to missing values are included in the training set, hence more reliable information but not necessarily the most relevant one (i.e., most important features).  Therefore, our framework includes this percentage threshold as a hyperparameter that needs to be studied and tuned according to the data at hand.

Finally, the training samples are further split into two groups. The first group consists of samples without any missing value, whereas in the second group, each sample has at least one missing value. Hence, the size of the initial data without any missing value changes with the pruning threshold selection.

\subsection{Data Handler: Feature Selection, Imputation, and Inlier Detection}
\label{sec:pipe1}

\begin{figure*}
    \centering
    \includegraphics[trim={0 0.3cm 0 0.0cm},clip,scale=0.8]{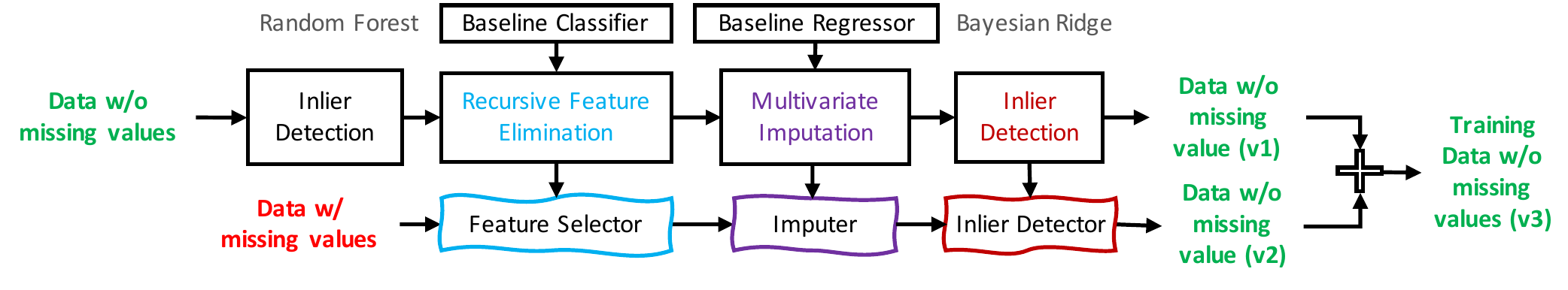}
    \caption{Data handler of proposed framework}
    \label{fig:framework1}
    \vspace{-0.5cm}
\end{figure*}

Fig. \ref{fig:framework1} illustrates the building blocks of the proposed data handler that aims at creating a dataset without any outliers and missing values while including only significant features. 
The training samples without missing values obtained from the preprocessing stage are used in this stage of our framework to build the baseline models for feature selection, data imputation, and inlier detection to be, then, applied to the training samples with missing values. 

The first step is to exclude outliers in the training data. We use Isolation Forest \cite{liu2008isolation}, an anomaly detection algorithm, to find the outliers. After removing the outliers, we use Recursive Feature Elimination with Cross-Validation (RFECV) to automatically find the most significant features. We utilize the Random Forest algorithm as the baseline classifier of the RFECV algorithm.  After training RFECV, a feature selection model is obtained which can be later used to eliminate unnecessary features of the wider input feature sets. Using the Isolation Forest prior to RFECV provides a more robust model for feature selection since RFECV can work better on clean data, without outliers. 

After selecting the most relevant features, we use a multivariate iterative imputer~\cite{buuren2010mice}, where missing values are imputed by modeling each feature with missing values as a function of other features in an iterated round-robin fashion \cite{pedregosa2011scikit}. We employ the Bayesian Ridge algorithm \cite{tipping2001sparse, mackay1992bayesian} as the baseline regressor of the multivariate imputer. This step provides us with an imputer model, which is later used to impute the missing data. Since the training dataset in this step does not have any missing values, this step achieves a reliable imputer, even with a limited number of samples. 

Having known the most relevant features, we propose to retrain the Isolation Forest model obtained in the first step of the data handler. The reason lies in the fact that we intend to use our inlier detector model on a selection of the input features. We propose to refit our inlier detection model after applying the imputer, since in a real scenario it is first required to impute the missing values, otherwise, the outlier detector fails to find the true outliers. 

Having obtained the models of feature selection, feature imputation, and inlier detection, we pass the second part of the training data containing missing values through these models, achieving training samples without unnecessary features, outliers, and missing values (v2). This already-cleaned part of training data can then be concatenated to the first part of the training data (v1) to create the final training dataset (v3).

Although using the initial clean and complete training data, v1, in the data handler provides us with the models of feature selection, imputation, and inlier detection, these models have been trained on a subset of the training dataset, i.e., those samples initially without any missing values. If this part of the training data is not sufficiently large, the overall framework is prone to overfitting. Therefore, one solution is to retrain the feature selector, multivariate imputer, and inlier detector with the complete and larger training dataset (v3) obtained from the \textit{initial} data handler. 
As a consequence of retraining the machine learning models within the data handler, new models for feature selection, inlier detection, and data imputation are attained, which can be later used at inference time. We refer to these models as \textit{retrained} data handler models. 

\subsection{Classifier}
To find the best classifier, we perform a grid search with cross-validation for several classical machine learning algorithms, including Linear Discriminant Analysis, Support Vector Classifier, Random Forest, and eXtreme Gradient Boosting (XGB) classifier similar to \cite{momeni2019real}. Although we only consider five of the well-known machine learning classifiers, our approach is not limited to incorporating these classifiers and any arbitrary machine learning algorithm can be employed within our proposed framework. 
In the grid search, we consider the most important hyperparameters of these algorithms. 
We found XGB able to provide a statistically higher CV score than the others. Therefore, we use XGB as the main classifier of the proposed framework.

\section{EXPERIMENTAL SETUP}
To assess our proposed technique, we build our framework in Scikit-Learn \cite{scikit-learn}. Then, we test it with experimental data from \cite{Rodrigues2020LocomotionChallenges} on a single core of a 32 AMD EPYC Processor with a maximum frequency of 2 GHz, a 500 GB main memory, and an 8 MB Last Level Cache (LLC). 

\subsection{Stress Database: Experiment Protocol}\label{sec:experiment}

95 participants (male, $Age_{mean}=20.43$, $Age_{std}=2.17$) are divided into two groups performing either a control or a stress task in a virtual reality (VR) environment lasting 10 minutes each one \cite{Rodrigues2020LocomotionChallenges}. The physiological signals are recorded using the Biopac BioNomadix System.  
The stress experiment is approved by the Cantonal Ethics Committee of Vaud, Switzerland (2017-00449).
The stress task exposed participants to an uncontrollable social-evaluative task and timed problem solving with negative feedback in a challenge in VR. 
Here, participants were immersed in an empty room with tiled flooring, in which they could move around while mental arithmetic questions appeared briefly in the heads-up display (HUD). 
Incorrect responses caused a tile on the floor to break and disappear, leaving an open hole where participants could fall into. Performance was continuously compared to a faux average performance from other participants (63\% of correct responses; being also shown in the HUD) and the difficulty (response time limit) was titrated to keep performance below this average. 
The control task consisted of equivalent conditions but without the stressful elements of the stress task. Participants were still standing up and allowed to walk while being immersed in a VR nature setting. 
\vspace{-0.1cm}
\subsection{Evaluation of the Proposed Framework }
To evaluate the utility of our framework to enhance the training data we compare the use of the data handler trained only by data without missing values against the two-step process of retraining our data handler on the complete training data that includes both the initial clean and treated data with the first data handler, see Fig. \ref{fig:overall_framework}. In particular, we compare how it behaves with respect to the different thresholds of the feature pruning, hence how robust our data handler is when the training data have more missing values.  

Moreover, we compare our framework with traditional techniques for handling missing data and outliers. The most conventional techniques for replacing the missing values are 1) to fill the gaps with the mean value of valid samples (i.e., those with neither missing values nor outliers) computed from the training set and 2) to replace the missing value with the value of the last valid sample. In the first data imputation technique, the mean value of each feature obtained from the training dataset is used to fill the missing values.
In addition, we also consider in our comparison the very basic yet commonly used alternative to handle missing values, where all the missing values are simply ignored \cite{bai2020fatigue}, discarding entire rows containing missing values. However, this comes at the price of losing valuable data.
Finally, the most common, yet simplistic, approach to consider the outliers is removing any values lying in a distance beyond 3 times of standard deviation from the mean value. We also assess this technique. 

The same signal processing as the one explained in Section \ref{sec:preprocesssing} is applied to the data prior to using these traditional techniques. Also, to have a fair comparison, we apply RFECV for feature selection followed by performing a grid search over the hyperparameters of the classifier (XGB) with cross-validation.
Finally, we remove any features that more than $50\%$ of their values in the training dataset are missing.

\section{EXPERIMENTAL RESULTS AND DISCUSSION}
\label{sec:results}

\subsection{Feature Pruning Threshold Selection}
\label{sec:threshold}
One of the steps taken in our framework in the preprocessing stage is to prune the features, i.e., to discard those features that more than a particular percentage of their samples in the training set are missing. 

\begin{figure}
    \centering
    \includegraphics[trim={0.0cm 0.250cm 0 0.0cm},clip, width=\columnwidth]{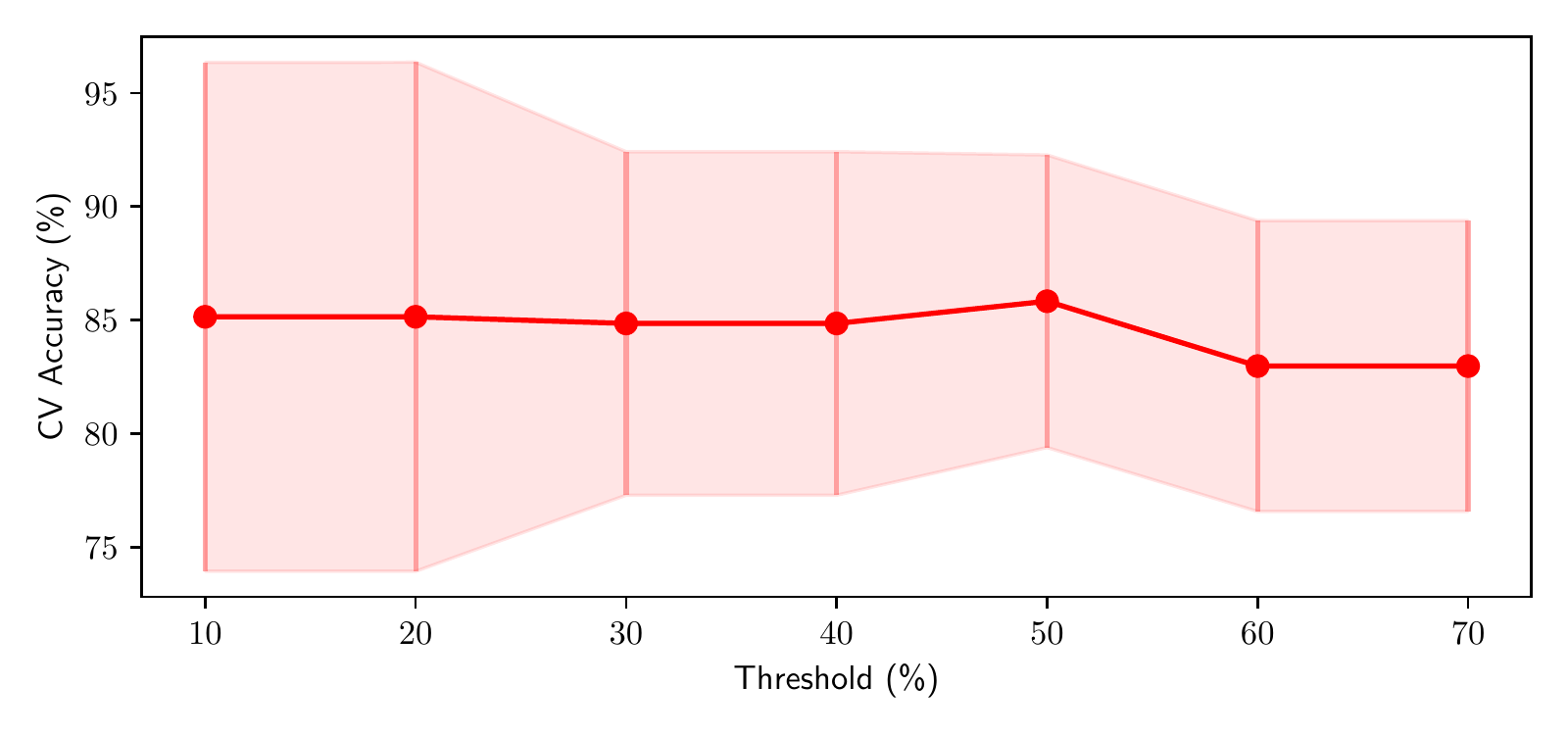}
    \caption{Impact of threshold for feature pruning cross validation results}
    \label{fig:threshold}
    \vspace{-0.25cm}
\end{figure}

Fig. \ref{fig:threshold} shows how this threshold affects the mean and standard deviation of CV accuracy. As shown in the figure, the CV mean accuracy of all threshold values range from $83.0\%$ to $86.8\%$ (solid line with markers) with a standard deviation from $6.4\%$ to $11.2\%$, indicating the robustness of the proposed framework. It is important to note that these results are highly dependent on the dataset used since the affected features with missing values vary with the sensor used and activities performed.

Moreover, the choice of threshold value affects the number of selected features and overhead of the framework. TABLE \ref{tab:threshold} shows the number of features after feature pruning, training samples without missing values after feature pruning used in the initial data handler, number of features after the retrained data handler used for training the classifier, and overhead of the whole framework at inference. As shown by TABLE \ref{tab:threshold}, the execution time proportionally increases with the number of features. In fact, the greater the number of features selected, the more complex would be the model of the imputer, inlier detector, and the final classifier.   

We argue that either of the thresholds evaluated in this work can be considered as the chosen value for the pruning step, depending on the user's intent and the application targeted. On one hand, if the main purpose is to attain a low-overhead solution, a threshold value of $10\%$ to $20\%$ is a proper choice as it can achieve an acceptable score of $85.15\%$ with a minimum runtime overhead of only $0.8$ ms. On the other hand, a threshold of 50\% provides slightly higher accuracy, i.e., 86.83\% at the cost of larger runtime overhead ($1.8ms$). The standard deviation of accuracy in cross-validation is, nonetheless, quite large for smaller thresholds. In this work, we assume $50\%$ as the threshold of missing data in the feature pruning step, as it provides the highest CV mean accuracy and the lowest standard deviation.

\begin{table}[t]
\centering
\caption{Impact of different thresholds for feature pruning}
\label{tab:threshold}
\begin{tabular}{|c|c|c|c|c|c|c|}
\hline
\textbf{Threshold (\%)}                                                                             & 10   & 20   & 30   & 40   & 50   & 60  \\ \hline
\textbf{\#Features (After Feat. Prun.)}                                                             & 66   & 66   & 80   & 80   & 86   & 87  \\ \hline
\textbf{\begin{tabular}[c]{@{}c@{}}\#Samples w/o missing values \\ (After Feat. Prun.)\end{tabular}} & 2475 & 2475 & 1925 & 1925 & 1498 & 675 \\ \hline
\textbf{\#Features (Final)}                                                                         & 21   & 21   & 27   & 27   & 38   & 30  \\ \hline
\textbf{Overhead (ms)}                                                                              & 0.8  & 0.8  & 1.5  & 1.5  & 1.8  & 1.6 \\ \hline
\end{tabular}
\vspace{-.5cm}
\end{table}

\subsection{Framework Analysis and Evaluation}
As explained in Section \ref{sec:pipe1} the models extracted from the initial data handler models can also be used at inference time. However, as aforementioned, the initial data handler is trained on only a part of the training data, v1. Therefore, the final feature selector, data imputer, and outlier detector from the retrained data handler using the whole clean dataset may outperform the models provided by the initial data handler. We test this hypothesis by deploying the proposed framework with the initial and retrained data handlers at different threshold values for feature pruning. 

\begin{figure}
    \centering
    \includegraphics[trim={0 0.5cm 0 0.0cm},clip,scale=0.6]{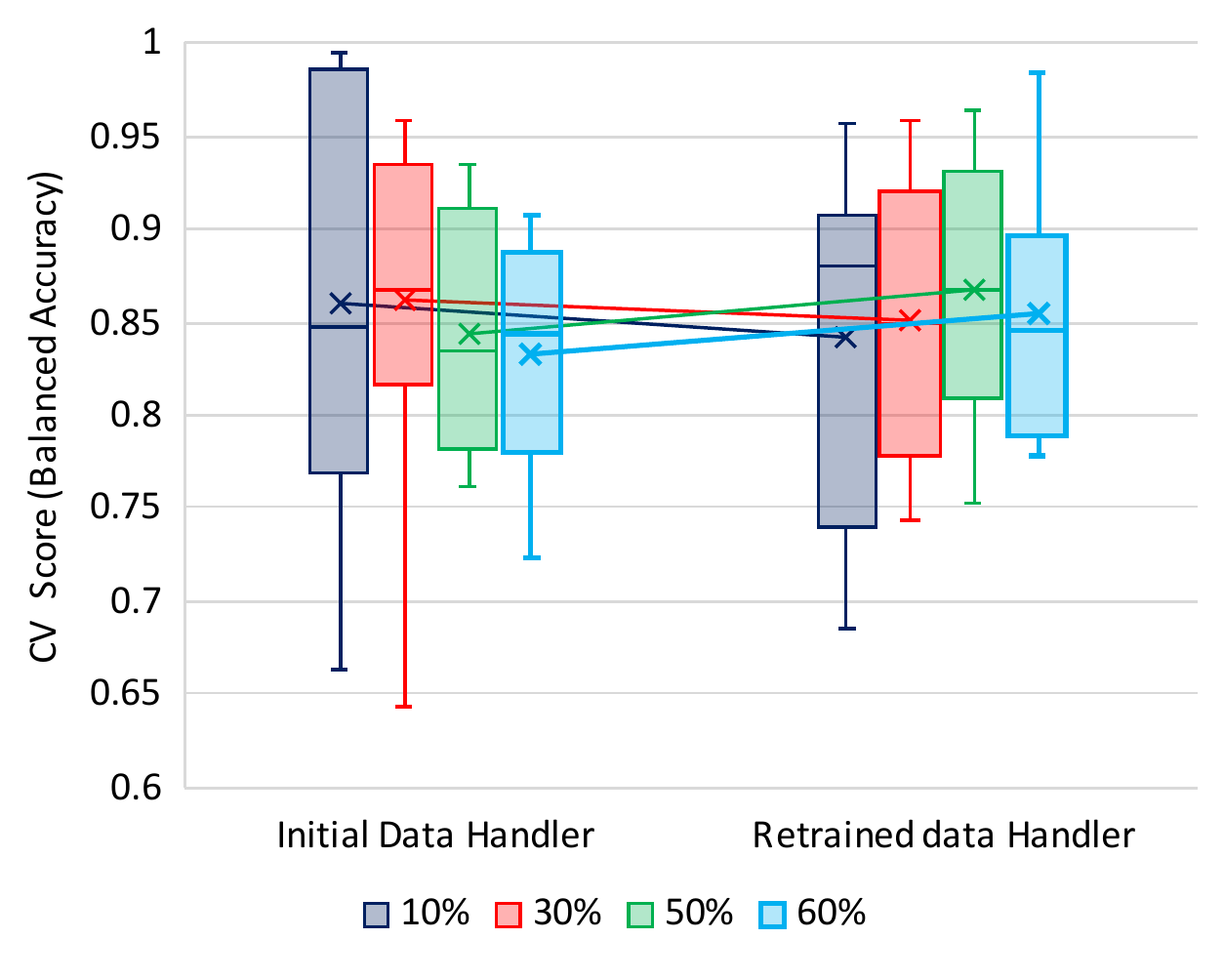}
    \caption{CV-accuracy obtained from models of the initial and retrained data handlers with respect to different threshold values}
    \label{fig:boxplot}
\end{figure}

\begin{figure}
    \centering
    \includegraphics[trim={0 0.8cm 0 0.25cm},clip, scale=0.6]{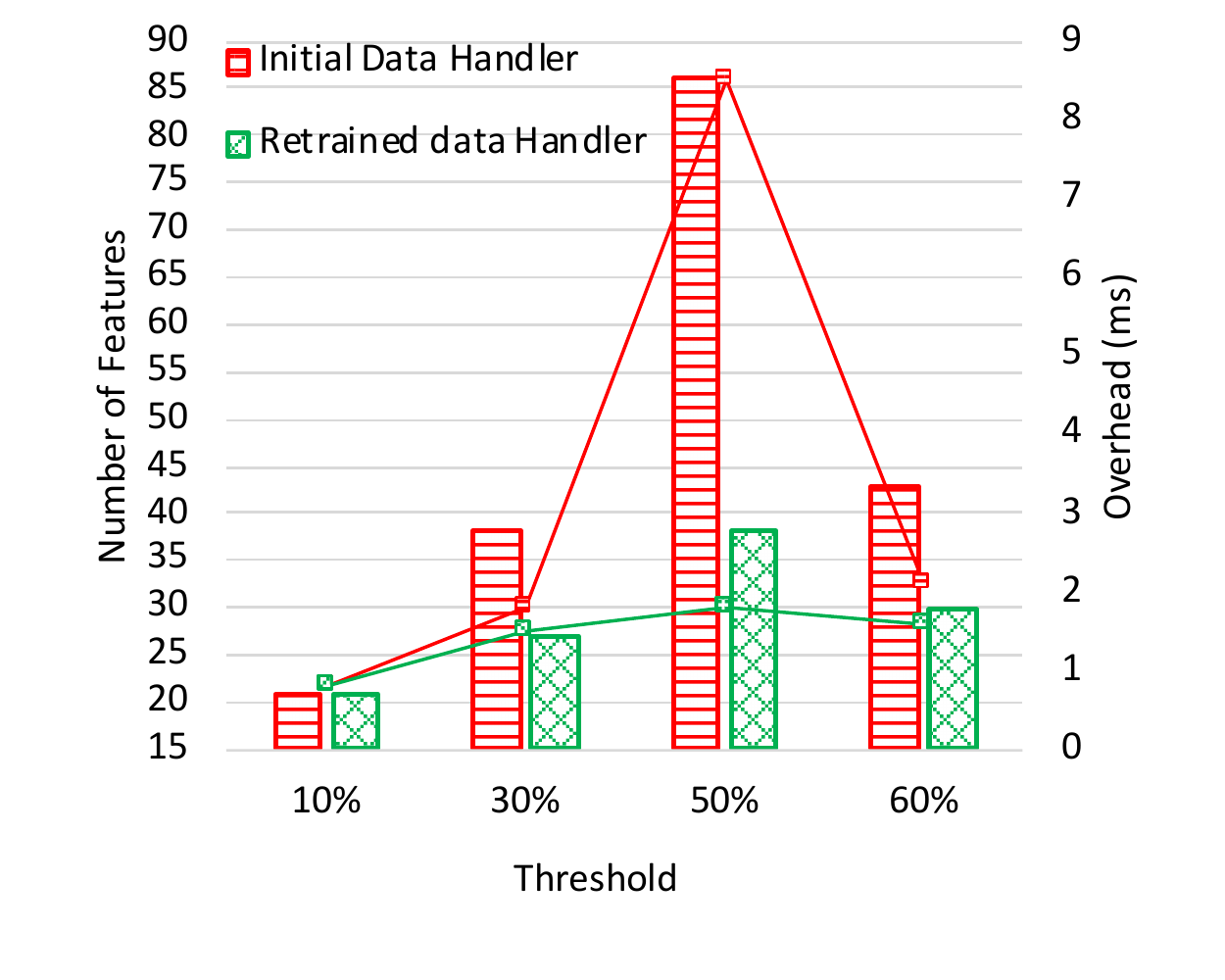}
    \caption{Number of features (bars) and overhead obtained by the initial and retrained data handlers with respect to different threshold values}
    \label{fig:barplot}
    \vspace{-0.5cm}
\end{figure}

Fig. \ref{fig:boxplot} shows the box plots of CV score achieved by models of the initial data handler and retrained data handler for different threshold values. With lower thresholds, the initial data handler attains a higher mean CV score than retrained data handler with a larger standard deviation. By increasing the threshold, retrained data handler provides not only a higher CV mean score, but also a lower standard deviation. As a consequence, inferring from the models of the retrained data handler brings about more robustness against missing data. Also, if applied to unseen testing data, with a feature pruning threshold of 50\%, the classification mean accuracy achieved through the retrained and initial data handlers are $78.8\%$ ($std=25.4\%$) and $76.5\%$ ($std=26.4$), respectively. 
Moreover, Fig. \ref{fig:barplot} depicts the number of features selected up to each of the approaches and the overall runtime overhead of the framework. According to this figure, the models of retrained data handler consistently use fewer features and, hence, come with lower complexity with different threshold values. Therefore, inferring from the models of retrained data handler results in reduced complexity while having statistically more accurate predictions.

\subsection{Comparison with Conventional Imputation Techniques}
\label{sec:comp_conventional}
In this section, we evaluate our approach against traditional techniques for coping with missing data and outliers. 
For this comparison, we assume the following approaches: filling the missing values with a mean value \textit{Mean value}; the last value\textit{Last Value} and ignoring all the missing values \textit{Drop NAN}, on the training dataset.

\begin{table}[t]
\centering
\caption{Comparison to conventional imputation techniques}
\label{tab:comp_conventional}
\begin{tabular}{l|c|c|c|c|}
\cline{2-5}
                                        & \begin{tabular}[c]{@{}c@{}}CV\\ (mean $\pm$ STD)\end{tabular} & \begin{tabular}[c]{@{}c@{}}Inf. (All)\\ (mean $\pm$ STD)\end{tabular} & \begin{tabular}[c]{@{}c@{}}Inf.\\ (Miss.)\end{tabular} & \#Feat.  \\ \hline
\multicolumn{1}{|l|}{Drop NaN}          & 81.6 $\pm$  12.0\%   & 74.4 $\pm$ 30.0\%     & N/A                                                           & 24          \\ \hline
\multicolumn{1}{|l|}{Mean value}        & 53.9 $\pm$ 13.5\%        & 50.2 $\pm$ 38.7\%                                                   & 51.1\%                                                          & 49          \\ \hline
\multicolumn{1}{|l|}{Last Value}        & 54.1 $\pm$  12.4\%      & 49.7 $\pm$ 38.2\%       & 50.5\%                                                          & 36          \\ \hline
\multicolumn{1}{|l|}{\textbf{ReLearn}} & \textbf{86.8 $\pm$ 6.4\%}  & \textbf{78.8 $\pm$ 25.4\%}                                          & \textbf{77.9\%}              & \textbf{38} \\ \hline
\end{tabular}
\vspace{-0.5cm}
\end{table}

TABLE \ref{tab:comp_conventional} compares our approach and these techniques with respect to the CV mean accuracy and standard deviation, inference mean accuracy and standard deviation among subjects on the unseen test data, mean accuracy of predictions on missing data at inference, and the number of selected features.  As shown by TABLE \ref{tab:comp_conventional}, by ignoring all missing data, the CV mean accuracy is close to that obtained by our framework, yet $4\%$ less. Besides, this technique results in a rather large standard deviation ($12.0\%$), thus, lower accuracy at inference. More importantly, regarding our database, the \textit{Drop NAN} method fails to provide any prediction for 387 samples out of 1131 samples. This is more than $34\%$ of the data to be predicted at inference. Even if we let only the features with less than $30\%$ of their values missing (instead of 50\% used in TABLE \ref{tab:comp_conventional}),  still no prediction can be made for $16\%$ of samples for unseen testing data. In this case, the CV mean accuracy and inference accuracy increase to $84\%$ and $77.4\%$, yet lower than the one achieved by the proposed framework for all threshold values discussed in Section \ref{sec:threshold}.

In contrast to the \textit{Drop NAN} method, using \textit{Mean Value} and \textit{Last value} techniques reduce overfitting through data imputation and, thus, increasing the training data, at the cost of lower CV and inference accuracy. 
In addition, despite the fact that predictions on missing data are made available by these two techniques, neither of them can reach the same accuracy as that provided by the proposed framework. The poor classification accuracy obtained through these two methods is mainly due to the large number of missing samples in the training and testing datasets, which necessitates a more complicated solution rather than these simple imputation techniques.


\section{CONCLUSION}
In this paper, we have proposed ReLearn, a new robust machine learning framework for stress detection from biomarkers extracted from multimodal physiological signals that effectively addresses missing data and outliers both at training and inference phases. Our framework enables efficiently increasing and cleaning the training data. Thus, it provides a more accurate and generalizable classification. Moreover, our framework allows classifying all samples at inference, including missing ones. In particular,  according to our experiments in a large stress database, while by discarding all missing data as a simplistic yet common approach, no prediction can be made for 34\% of the data at inference, our approach is able to achieve very accurate predictions, as high as 78\%, for missing samples.
Furthermore, we have shown that our approach facilitates the use of features that are usually discarded due to missing values, despite containing significant information of the physiological stress response. Thus, our approach achieves a cross-validation and inference accuracy of 86.8\% and 78.8\%, respectively, even if up to 50\% of samples within the features are missing.   

\bibliographystyle{IEEEtran}
\bibliography{References}

\end{document}